%% file: main.tex
\documentclass[10pt,twocolumn,letterpaper]{article}

\usepackage{iccv}
\usepackage{times}
\usepackage{epsfig}
\usepackage{graphicx}
\usepackage{amsmath}
\usepackage{amssymb}

\usepackage{booktabs}
\usepackage{multirow}
\usepackage{bbding}
\usepackage{caption}
\usepackage[pagebackref=true,breaklinks=true,letterpaper=true,colorlinks,bookmarks=false]{hyperref}

\usepackage[capitalize]{cleveref}
\crefname{section}{Sec.}{Secs.}
\Crefname{section}{Section}{Sections}
\Crefname{table}{Table}{Tables}
\crefname{table}{Tab.}{Tabs.}
\Crefname{figure}{Figure}{Figures}
\crefname{figure}{Fig.}{Figs.}
\crefname{equation}{Equ.}{Equs.}

\iccvfinalcopy % *** Uncomment this line for the final submission

\def\iccvPaperID{5733} % *** Enter the ICCV Paper ID here

\ificcvfinal\fi

\begin{document}

%%%%%%%%% TITLE
\title{Reconstructing Groups of People with Hypergraph Relational Reasoning}

\author{Buzhen Huang$^{1}$\hspace{5mm} Jingyi Ju$^{1}$\hspace{5mm} Zhihao Li$^{2}$\hspace{5mm} Yangang Wang$^{1}$\footnotemark[1]\\%
\\
$^1$Southeast University, China\\
$^2$Huawei Noah’s Ark Lab\\
}

% \maketitle
\twocolumn[{%
\renewcommand\twocolumn[2][]{#1}%
\maketitle
\thispagestyle{empty}
% \vspace{-10mm}
\begin{center}
   \centering
   \vspace{-7mm}
   \includegraphics[width=0.95\textwidth]{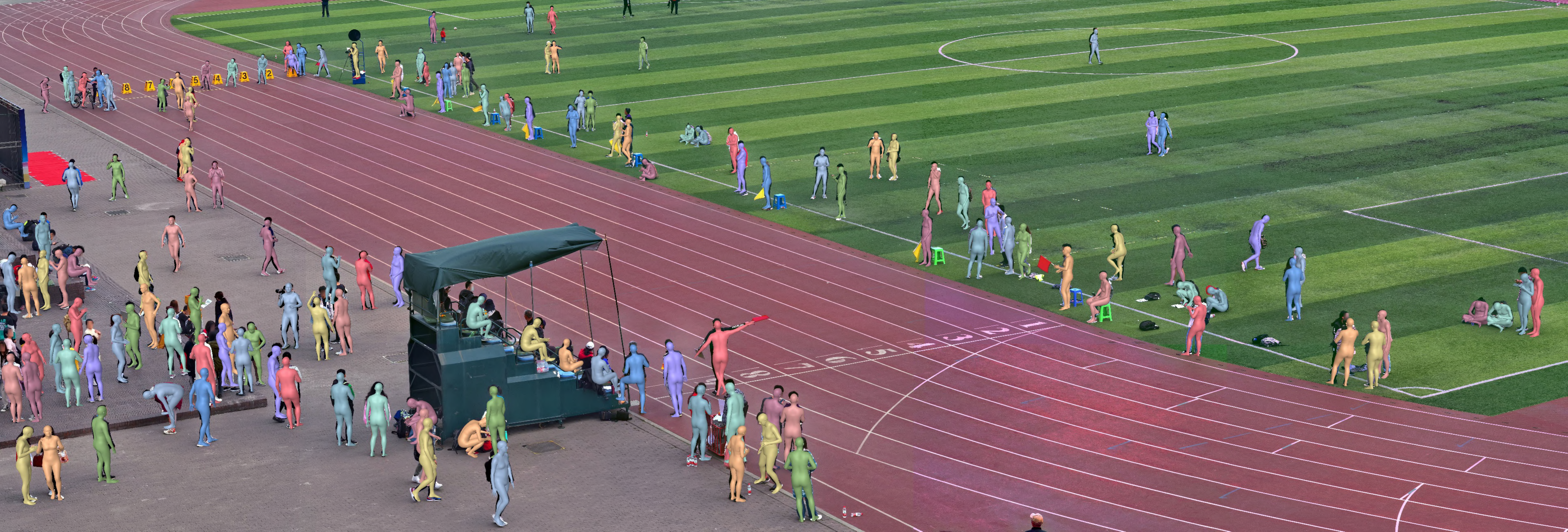}
   \vspace{-2mm}
   \captionof{figure}{We exploit human collectiveness and correlation in crowds to improve human mesh recovery in large-scale crowded scenes (more than 50 people).} 
   \label{fig:teaser}
\end{center}
% \vspace{-3mm}
}]
% Remove page # from the first page of camera-ready.
\ificcvfinal\thispagestyle{empty}\fi

\renewcommand{\thefootnote}{\fnsymbol{footnote}}
\footnotetext[1]{Corresponding author. E-mail: yangangwang@seu.edu.cn. All the authors from Southeast University are affiliated with the Key Laboratory of Measurement and Control of Complex Systems of Engineering, Ministry of Education, Nanjing, China. This work was supported in part by the National Natural Science Foundation of China (No. 62076061), the Natural Science Foundation of Jiangsu Province (No. BK20220127).}

%%%%%%%%% ABSTRACT
\begin{abstract}
   Due to the mutual occlusion, severe scale variation, and complex spatial distribution, the current multi-person mesh recovery methods cannot produce accurate absolute body poses and shapes in large-scale crowded scenes. To address the obstacles, we fully exploit crowd features for reconstructing groups of people from a monocular image. A novel hypergraph relational reasoning network is proposed to formulate the complex and high-order relation correlations among individuals and groups in the crowd. We first extract compact human features and location information from the original high-resolution image. By conducting the relational reasoning on the extracted individual features, the underlying crowd collectiveness and interaction relationship can provide additional group information for the reconstruction. Finally, the updated individual features and the localization information are used to regress human meshes in camera coordinates. To facilitate the network training, we further build pseudo ground-truth on two crowd datasets, which may also promote future research on pose estimation and human behavior understanding in crowded scenes. The experimental results show that our approach outperforms other baseline methods both in crowded and common scenarios. The code and datasets are publicly available at \url{https://github.com/boycehbz/GroupRec}.
\end{abstract}

\input{introduction.tex}

\input{relatedwork.tex}

\input{method.tex}

\input{dataset.tex}
\input{experiments.tex}

\input{conclusion.tex}

\noindent\textbf{Acknowledgments.}
% \section*{}
The authors would like to thank the anonymous reviewers for their valuable comments. They also thank Yizhu Li for helpful discussions. 

{\small
\bibliographystyle{ieee_fullname}
\bibliography{egbib}
}

\end{document}

%% file: introduction.tex
% \vspace{-8mm}
\section{Introduction}\label{sec:introduction}
Although immense progress has been made in monocular multi-person human mesh recovery in recent years, the existing methods still cannot accurately reconstruct groups of people from large-scale crowded scenes. The top-down formulation~\cite{choi2022learning,khirodkar2022occluded,huang2022pose2uv,jiang2020coherent} iteratively predicts each individual from tightly cropped image patches, which discards the interaction relationships and location information in the original camera coordinates. Alternatively, the bottom-up formulation~\cite{sun2022putting,zhang2021body,sun2021monocular,zanfir2018deep} parses inter-person interactions with global pixel-level cues and enables its impressive performance on occluded cases. However, bottom-up methods always fail in large-scale scenes like~\cref{fig:teaser} since they require downsampling images to low-resolution ~(\eg, 512$\times$512) to satisfy computational constraints.

Recently, a few works have attempted to estimate human poses in large-scale crowded scenes. Several techniques like composite fields~\cite{kreiss2019pifpaf} and occlusion augmentation~\cite{golda2019human,huang2022object} in 2D pose estimation are proposed for addressing the low-resolution inputs. PandaNet~\cite{benzine2020pandanet} further lifts the root-relative 3D poses from the 2D detections with an anchor-based representation. Nevertheless, these works cannot be used to reconstruct absolute body meshes in camera coordinates due to the inherent coupling between depth and body shape. In addition, the challenges of a huge number of people, severe mutual occlusions, and complex spatial distribution make the problem far from being solved.

\begin{figure}
    \begin{center}
    \includegraphics[width=1.0\linewidth]{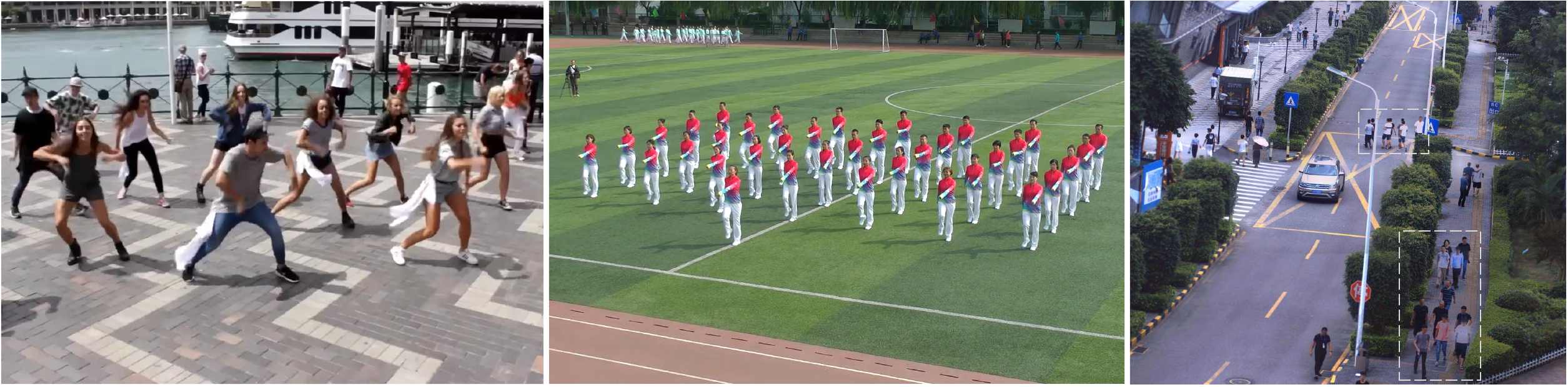}
    \end{center}
    \vspace{-6mm}
    \caption{Collective motions are common in human crowd.}
\label{fig:collectiveness}
\vspace{-7mm}
\end{figure}

Different from a few people or single-person cases, it is very common that the crowd in large-scale scenes show significant interactive and collective motions~\cite{zhou2013measuring,mei2019measuring}. As shown in \cref{fig:collectiveness}, the individuals in the same group even have similar poses. Based on this observation, \textbf{our key-idea is to fully exploit the collectiveness and social interaction in crowds, and promote human mesh recovery in large-scale scenes with a relational reasoning.}

However, the idea faces two technical obstacles. First, without a compact representation, the limited hardware memory cannot afford the relational reasoning for a large number of people. Second, the existing networks can hardly formulate the complex and high-order correlation among different individuals and groups in the crowd. To address the obstacles, we propose a multiscale hypergraph to represent the individuals and groups in different scales, and discard the redundant image features in the relational reasoning. Specifically, we first detect bounding-boxes~\cite{akyon2022sahi,yolox2021} and extract valid human features in the original image. Different from previous top-down methods, we also record the bounding-box information, which preserves the vital global location cues~\cite{cliff} to regress humans in absolute camera coordinates. The compact features and corresponding bounding-boxes depict expressive and high-resolution human information in the crowd image. Then, a multiscale hypergraph network is constructed for the relational reasoning. Based on the hypergraph structure~\cite{feng2019hypergraph}, we represent the individuals with hypergraph nodes, and the nodes on the same hyperedge are regarded as a group. Since human groups in a crowd image have unordered structure, the connection relationships of hyperedges cannot be defined with hand-crafted adjacency matrix like previous graph-based pose estimation methods~\cite{cai2019exploiting,choi2020pose2mesh}. We thus introduce a differentiable optimization to infer the graph topology, and then assign the individuals with high human feature similarity to the same group. Subsequently, we initialize the nodes with individual human features, and the features for different individuals and groups can pass through the hypergraph via node-to-hyperedge and hyperedge-to-node phases. After the relational reasoning, the updated individual features with group information in the nodes can be utilized to regress the groups of people with absolute positions. In addition, since no existing 3D human dataset is captured in real large-scale scenes, we further build pseudo ground-truth on Panda~\cite{wang2020panda} and CrowdPose~\cite{li2019crowdpose} to relieve the domain gap for synthetic data. The datasets may promote future research on pose estimation and human behavior understanding in large-scale scenes. The main contributions of this work are summarized as follows.

\begin{itemize}
    \vspace{-2mm}
    \item We reconstruct crowds from single color images and verify that crowds can provide essential knowledge for multi-person mesh recovery.
    \vspace{-2mm}
    \item We propose a hypergraph relational reasoning network to formulate correlations among individuals and groups, which exploits crowd collectiveness and social interaction to improve human mesh recovery in crowded scenes.
    \vspace{-2mm}
    \item We build pseudo ground-truth on 2 crowd datasets to promote the research on pose estimation and human behavior understanding in large-scale crowded scenes.
\end{itemize}

%% file: relatedwork.tex
\section{Related Work}\label{sec:relatedwork}
\noindent\textbf{2D multi-person pose estimation.}
2D multi-person pose estimation explicitly considers person-person interactions and occlusions~\cite{zuo2023reconstructing}, which can be generally divided into two categories: top-down and bottom-up methods. The top-down strategy~\cite{chen2018cascaded,papandreou2017towards,he2017mask,guler2018densepose} iteratively performs pose estimation on each individual in the image. The method achieves high accuracy, but the detection errors in crowded scenes may result in poor performance~\cite{fang2017rmpe}. The bottom-up strategy~\cite{pishchulin2016deepcut,insafutdinov2016deepercut,iqbal2016multi,kocabas2018multiposenet} distinguishes the body parts of different people simultaneously and produces more robust results in interactive cases. Some representative grouping approaches like Part Affinity Field~\cite{cao2019openpose}, Associative Embedding~\cite{newell2017associative}, and mid-range offset fields~\cite{papandreou2018personlab} are introduced to assemble limbs. However, directly applying these methods in large-scale images~(\eg, gigapixel-level~\cite{wang2020panda} and surveillance~\cite{cormier2021interactive} video) may fail to obtain satisfactory results. The top-down models discard the interactive cues in the original image from the very beginning, while the bottom-up models confront severe scale variations. Only a few works attempt to address the challenges of low resolution and mutual occlusions in larger-scale crowd images with synthesis data~\cite{fabbri2018learning,golda2019human}, composite fields~\cite{kreiss2019pifpaf} and association mechanism~\cite{li2019crowdpose}. Nevertheless, all of them do not utilize relationships among individuals like pose similarity and crowd collectiveness~\cite{zhou2013measuring} in the pose estimation.

\noindent\textbf{3D multi-person pose and shape reconstruction.} 
3D multi-person pose estimation~\cite{moon2019camera,rogez2019lcr,mehta2018single,fabbri2020compressed,reddy2021tessetrack} directly regresses joint positions from images, which faces inherent depth ambiguity. To obtain correct depth order in camera-centric coordinate, compressed volumetric heatmap~\cite{fabbri2020compressed}, ordinal relation~\cite{wang2020hmor}, camera prior knowledge~\cite{moon2019camera,mehta2019xnect,huang2021dynamic}, and root depth map~\cite{zhen2020smap,jin2022single,liu2022explicit,cheng2022dual,wang2022distribution} are proposed for absolute pose prediction. Due to the inherent shape-depth coupling, the multi-person shape reconstruction is more ambiguous than pure pose estimation. The absolute position may not be available~\cite{sun2021monocular,choi2022learning}, or the estimation requires additional ground plane constraints~\cite{zanfir2018monocular,ugrinovic2021body}. Some works~\cite{zanfir2018deep,huang2022pose2uv} can regress translations with 2D poses and focal length, but the strategy is strongly affected by the accuracy of 2D poses and predicted body shapes. Other works utilize 6D pose estimation~\cite{mustafa2021multi}, point-based representation~\cite{zhang2021body}, bird’s-eye-view-based representation~\cite{sun2022putting}, and depth ordering-aware loss~\cite{jiang2020coherent,khirodkar2022occluded} to address the obstacles. However, these solutions in multi-person mesh recovery cannot easily be applied to large-scale scenes due to the low resolution and computational constraints. Recently, Crowd3D~\cite{wen2023crowd3d} estimates SMPL maps for cropped patches, and relies on a calibrated ground plane to combine the results in global coordinates. In this work, we incorporate group features and location information~\cite{cliff} in the network inference, and supervise 3D humans in the original camera coordinate system. Unlike the recent relation-aware work~\cite{kimmulti}, our method explicitly considers the group-wise relations with crowd collectiveness, producing more accurate results for the occluded people in crowded scenes.

\noindent\textbf{Crowd analysis.}
Crowd analysis has broad applications in visual surveillance, social behavior understanding, density measurement, and abnormal activity detection. The earliest work in crowd analysis is found in crowd counting~\cite{grant2017crowd}, which counts individuals~\cite{zheng2022progressive} or approximates the density of the crowd~\cite{liu2022leveraging,shu2022crowd}. Other works further exploit the interactions within crowds for behavior understanding. The underlying relationships among people are utilized in activity recognition~\cite{ibrahim2018hierarchical}, dominant motion extraction~\cite{cheriyadat2008detecting}, and trajectory prediction~\cite{xu2018encoding,xu2022groupnet}. Nevertheless, previous works in crowd analysis always adopt simplified models~(\eg, particle system~\cite{ali2007lagrangian}) to represent the crowd, which discards a lot of human details. In this work, we recover multi-person meshes with absolute positions from a monocular crowd image. To address the occlusions and interactions in crowds, we exploit the crowd collectiveness~\cite{zhou2013measuring,mei2019measuring}, which indicates the degree of individuals acting as a union in collective motion, and formulate the complex and high-order relation correlation with a hypergraph relational reasoning network. The reconstructed high-fidelity 3D human provides more information to describe the crowd, which may promote future research on human behavior understanding.

%% file: method.tex
\section{Method}\label{sec:method}

\begin{figure*}
    \begin{center}
    \includegraphics[width=1.0\linewidth]{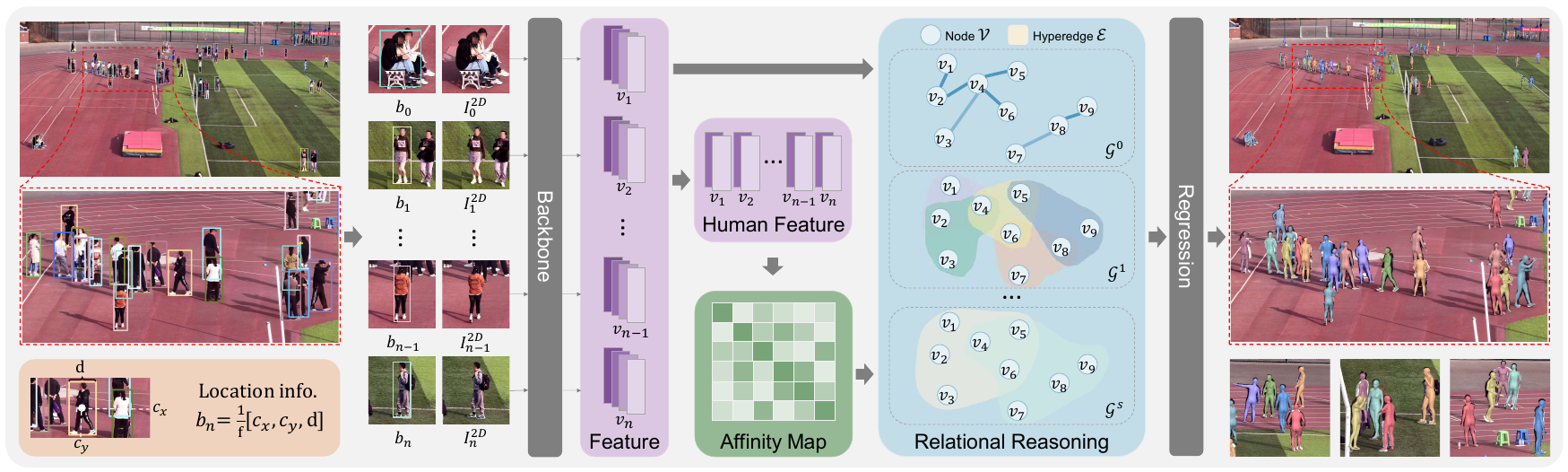}
    \end{center}
    \vspace{-6mm}
    \caption{Overview of our method. We first extract location information $b$ and high-resolution human features $q$ from the original crowd image. Then, we infer the graph topology according to crowd collectiveness, and then represent the individuals and groups in the crowd with nodes $\mathcal{V}$ and hyperedges $\mathcal{E}$ of multiscale hypergraphs $\mathcal{G}$. By conducting the hypergraph relational reasoning, we exploit the group features to provide additional cues to regress a crowd of people in camera coordinates.}
\label{fig:pipeline}
\vspace{-6mm}
\end{figure*}

Our method reconstructs groups of people from a large-scale crowd image. The compact human features are first extracted for each individual~(\cref{sec:extraction}). Then, a hypergraph relational reasoning network is constructed to fully exploit the collectiveness and interaction relationship among individuals and groups in the crowd~(\cref{sec:reasoning}). Finally, the group features can compensate for insufficient individual information to regress the 3D crowd with accurate body poses and shapes~(\cref{sec:regression}).

\subsection{Preliminaries}\label{sec:Preliminaries}
\textbf{Representation.} We adopt SMPL model~\cite{loper2015smpl} with 6D rotation representation~\cite{zhou2019continuity} to represent the 3D human. The model consists of pose $\theta \in \mathbb{R}^{144}$, shape $\beta \in \mathbb{R}^{10}$ and translation $t \in \mathbb{R}^{3}$ parameters. Finally, the output of our network for $N$ people are $\left\{\theta_1, \beta_1, t_1, \cdots, \theta_N, \beta_N, t_N\right\} \in \mathbb{R}^{N \times 157}$.

\textbf{Hypergraph neural networks~(HGNN).} HGNN can formulate complex and high-order data correlation with high efficiency through its hypergraph structure~\cite{feng2019hypergraph}. It can be defined as $\mathcal{G}=\left(\mathcal{V}, \mathcal{E}\right)$, where $\mathcal{V}$ and $\mathcal{E}$ are the set of nodes and hyperedges. Different from the simple graph, a hyperedge connects two or more nodes, where the connection relationships are defined by an adjacency matrix $\mathcal{H}$. In this work, we adopt the nodes and hyperedges to represent individuals and groups, and exploit the crowd collectiveness and interactions with a hypergraph relational reasoning.

\subsection{Individual feature extraction}\label{sec:extraction}
We first extract human features from the crowd image for the relational reasoning. Since each person occupies only a small proportion of pixels in a large-scale image~\cite{wang2020panda}, it is nontrivial to extract valid and high-quality human features from such inputs. The previous bottom-up methods directly rescale the original image for network input, which results in extremely low resolution and then leads to poor reconstruction performance. Consequently, we predict bounding-boxes for each human and then extract valid image features from the original image as input.

Specifically, we first predict all bounding-boxes~\cite{yolox2021} from the large image. To preserve the localization information, we transform the box coordinates to $b_{n}= \frac{1}{f} \left[c_x, c_y, d\right]$, where $n \in[1, \ldots, N]$. $N$ is the number of people in the image. $(c_x, c_y)$ is the box location relative to the original image center, and $d$ is its size. $f$ is the focal length of the original image. With the predicted bounding-boxes, the image patches for all people $\mathcal{I}^{2 D}=\left\{I_n\right\}$ on the original image can be cropped. We then extract the high-resolution human image features $q_n \in \mathbb{R}^m$ from the image patch with a backbone network.

\subsection{Hypergraph relational reasoning}\label{sec:reasoning}
Due to the occlusions and depth ambiguity, the individual features are insufficient to regress an accurate 3D human in crowded scenes. However, crowds always show significant collective and interactive motions. The group information can provide additional knowledge for the reconstruction. The core of our work is to exploit the collectiveness and interaction relationship in crowds for multi-person mesh recovery. We propose a novel multiscale hypergraph network to formulate complex correlations among individuals and groups. Mathematically, the individuals are denoted as nodes $\mathcal{V}=\left\{v_1, v_2, \cdots, v_N\right\}$, and the groups at scale $s$ are represented with hyperedges $\mathcal{E}^{(s)}=\left\{e_1^{(s)}, e_2^{(s)}, \cdots, e_{M_s}^{(s)}\right\}$. The nodes on the same hyperedge belong to the same group, and a larger $s$ indicates a larger group size. Hence, the multiscale hypergraph are defined as $\mathcal{G} = \left\{\mathcal{G}^{(0)}, \mathcal{G}^{(1)}, \ldots \mathcal{G}^{(S)}\right\}$, where $\mathcal{G}^{(s)}=\left(\mathcal{V}, \mathcal{E}^{(s)}\right)$.

Previous graph-based relational reasoning methods~\cite{adeli2021tripod,li2020evolvegraph} only focus on modeling the pair-wise interaction, which ignores the group-wise correlations. In contrast, our hypergraph explicitly forms group structures to exploit human collectiveness and considers the group’s influence on individuals. Besides, the multiscale design also alleviates the over-smoothing in conventional graph networks~\cite{zhou2020towards}.

\textbf{Collectiveness based group inference.} To define the connection relationship of hyperedges, adjacency matrices $\mathcal{H}^{(s)} \in \mathbb{R}^{|\mathcal{V}| \times\left|\mathcal{E}^{(s)}\right|}$ are also required. They show the topology of hypergraph, where $\mathcal{H}_{i, j}^{(s)}=1$ if the $i$th node is included in the $j$th hyperedge, otherwise $\mathcal{H}_{i, j}^{(s)}=0$. Existing graph-based pose estimation~\cite{cai2019exploiting,choi2020pose2mesh} builds hand-crafted adjacency matrices with known human skeletal and symmetrical relationships. However, human groups are unordered structures without intuitive interpretations. Thus, we infer the topology with an implicit human pose similarity.

The people with similar human features $v_n= [q_n, b_n] \in \mathbb{R}^{m+3}$ are assigned to the same group. We first compute an affinity matrix $\mathcal{A} \in \mathbb{R}^{N \times N}$ based on the human feature correlation:
\begin{equation}
    \mathcal{A}_{i, j}=v_i^{\top} v_j /\left(\left\|v_i\right\|_2\left\|v_j\right\|_2\right).
\end{equation}
The element of $\mathcal{A}_{i, j}$ measures the pose similarity and spatial proximity between $i$th and $j$th individuals. For $\mathcal{G}^{(0)}$, we consider the common pair-wise relationship. The two nodes with the largest affinity scores will be connected, leading to adjacency matrix $\mathcal{H}^{(0)}$ and hyperedge $\mathcal{E}^{(0)}$. The other hypergraphs consider group-wise relationships. Assuming the group size at $s$th scale is $K^{(s)}$, we then find the $K^{(s)} \times K^{(s)}$ high-density submatrices in $\mathcal{A}$. The $K^{(s)}$ nodes in the group have the highest correlation, and we use a hyperedge $e_i^{(s)}$ at scale $s$ to represent the group. The hyperedge can be obtained with:
\begin{equation}
    \begin{aligned}
    &e_i^{(s)}=\arg \max _{\Omega \subseteq \mathcal{V}}\left\|\mathcal{A}_{\Omega, \Omega}\right\|_{1,1} \\
    &\text { s.t. }|\Omega|=K^{(s)}, v_i \in \Omega, i=1, \ldots, N,
    \end{aligned}
\end{equation}
where $\|\cdot\|_{1,1}$ is the sum of the absolute values of all elements. The optimization can be efficiently solved with a greedy algorithm. For each node $v_i$, we find other $K^{(s)} - 1$ nodes to form a group. Therefore, the hypergraph at scale $s$ has $N$ hyperedges. That is, the hypergraphs at different scales have the same number of nodes. Finally, we obtain all $\mathcal{H} = \left\{\mathcal{H}^{(0)}, \mathcal{H}^{(1)}, \ldots \mathcal{H}^{(S)}\right\}$ to construct the hypergraphs.

\noindent\textbf{Group message passing}
Once the graph topologies are constructed, we initialize the node with the individual features $v_n$. Different from simple graphs, we can directly exploit group-wise correlations of all group members with the hypergraph structure, and then use the group features to compensate for each individual. To achieve the group message passing, we design node-to-hyperedge and hyperedge-to-node phases. In the node-to-hyperedge phase, the individual features in the nodes are first aggregated to hyperedge to obtain group features. Then, the group features are used to update the corresponding individual in the hyperedge-to-node phase. We iteratively execute the two phases at different scales. Finally, the individual features at all scales are concatenated to decode the human pose parameters.

Specifically, the group features are obtained with the following function in the node-to-hyperedge phase:
\begin{equation}
    \mathbf{e}_i=c_i \mathcal{F}_{e}\left(\sum_{v_j \in e_i} \lambda_j v_j\right),
\end{equation}
where $\lambda_j=\mathcal{F}_{\lambda}\left(v_j, \sum_{v_m \in e_i} v_m\right)$, which denotes the contribution of the $j$th node to the $i$th group. $c_i$ is the group collectiveness factor:
\begin{equation}
    c_i= \sigma \left( \mathcal{F}_{c}\left(\sum_{v_j \in e_i} (v_j - \bar{v}_i)\right) \right) ,
\end{equation}
where $\sigma (\cdot)$ is a sigmoid function, and $\bar{v}_i$ is the average features of the $i$th group. $\mathcal{F}_{e}$, $\mathcal{F}_{\lambda}$, and $\mathcal{F}_{c}$ are learnable functions implemented by MLPs.

In the hyperedge-to-node phase, the aggregated group features on all associated hyperedges are used to update the individual features. 
\begin{equation}
    v_i = \mathcal{F}_{v}\left(v_i, \sum_{e_j \in \mathcal{E}_i} \mathbf{e}_j\right),
\end{equation}
where $\mathcal{E}_i=\left\{e_j \mid v_i \in e_j\right\}$ denotes the all hyperedges that associate with $v_i$. $\mathcal{F}_{v}$ is also implemented with MLPs. 

The node-to-hyperedge and hyperedge-to-node phases are simultaneously repeated for several times in all scales of hypergraphs. The final individual features on the nodes contain group features and interaction information, which promote more reasonable spatial distribution. Furthermore, the group with high pose similarity can provide additional gesture knowledge for the occluded person to infer a plausible 3D mesh.

\subsection{Human parameter regression}\label{sec:regression}
After the relational reasoning, the node features on different scales are concatenated with the bounding-box information to obtain the final individual representation $v^{\prime}_i = \left[v_i^{(0)}, v_i^{(1)}, \cdots, v_i^{(S)}, b_i\right]$, which is used to regress the pose $\theta$, shape $\beta$ and camera $[f_c, t_x, t_y]$ parameters. The predicted camera can be further transformed into absolute translation:
\begin{equation}
    t_X =t_x + \frac{2 c_x}{d f_c}, t_Y =t_y + \frac{2 c_y}{d f_c}, t_Z =\frac{2 f}{d f_c}, \label{equ:translation}
\end{equation}
where $t = [t_X, t_Y, t_Z]$ is the translation. More details on the transformation can refer to \cite{cliff}. Finally, the network output all SMPl parameters $\left\{\theta_1, \beta_1, t_1, \cdots, \theta_N, \beta_N, t_N\right\} \in \mathbb{R}^{N \times 157}$ for $N$ people.

\begin{figure*}
    \begin{center}
    \includegraphics[width=1.0\linewidth]{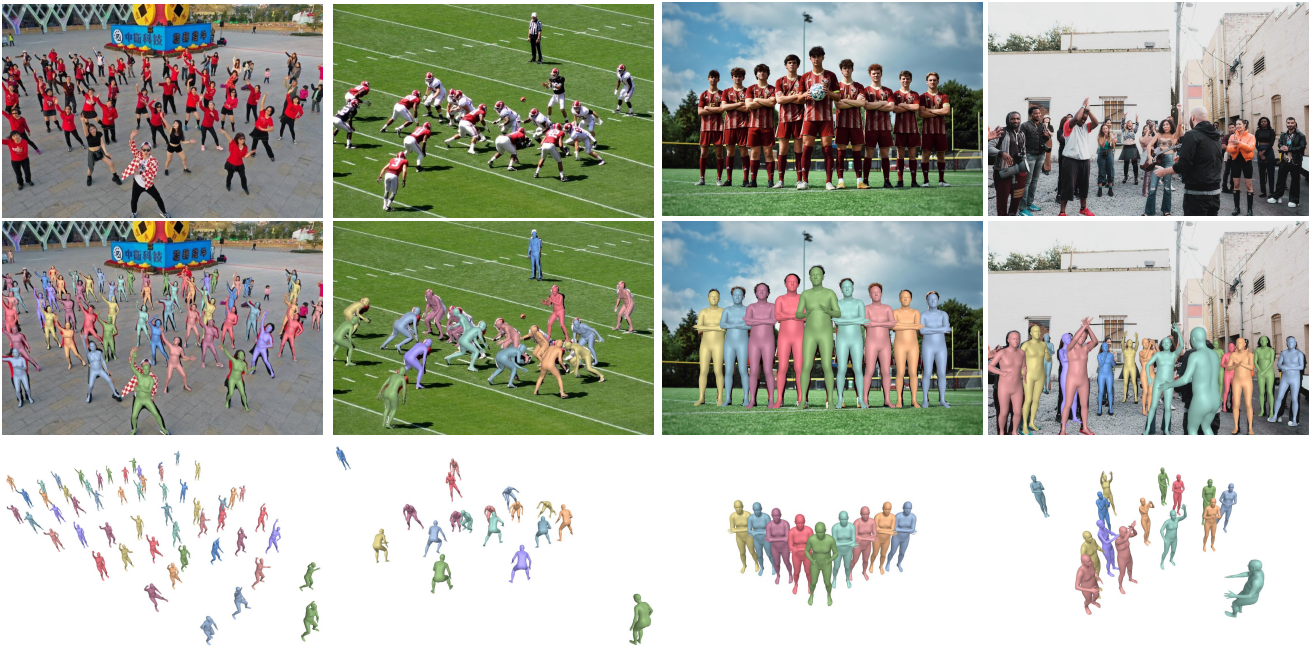}
    \end{center}
    \vspace{-8mm}
    \caption{Our method produces accurate body poses and reasonable spatial distribution on \textbf{Internet images}~\cite{Pexels}.}
\label{fig:qualitative_results}
\vspace{-6mm}
\end{figure*}

\subsection{Network training}\label{sec:training}
The network is trained in an end-to-end manner with the following loss function:
\begin{equation}
    \mathcal{L} = \lambda_{1} \mathcal{L}_{\text{reproj}} + \lambda_{2} \mathcal{L}_{\text{smpl}} + \lambda_{3} \mathcal{L}_{\text{joint}} + \lambda_{4} \mathcal{L}_{\text{crowd}},\label{equ:loss}
\end{equation}
where $\lambda_{1}=5.0$, $\lambda_{2}=5.0$, $\lambda_{3}=5.0$, and $\lambda_{4}=0.1$ are loss weights. With the transformation in \cref{equ:translation}, we can supervise the reprojection error in the original image, which enforces the network to regress reasonable absolute translations. Specifically, we add the translation $t$ to the SMPL 3D joint positions $J_{3D}$ and calculate the loss with following function:
\begin{equation}
    \mathcal{L}_{\text{reproj}} = \frac{1}{N} \sum_{n=1}^N \| \Pi\left(J_{3 D}^n+ t^n\right) - \hat{J_{2D}^n} \|_2^2,\label{equ:reproj}
\end{equation}
where $\Pi$ projects the 3D joints to 2D image with camera parameters, and $\hat{J_{2D}^n}$ is ground-truth 2D pose for $n$th person. The SMPL parameters and 3D joint positions are also used for supervision:
\begin{equation}
    \mathcal{L}_{\text{smpl}} = \frac{1}{N} \sum_{n=1}^N \| [\beta^n, \theta^n] - [\hat{\beta^n}, \hat{\theta^n}] \|_2^2.
\end{equation}
\begin{equation}
    \mathcal{L}_{\text{joint}} = \frac{1}{N} \sum_{n=1}^N \| J_{3D}^{n} - \hat{J_{3D}^{n}} \|_2^2.
\end{equation}
The $\hat{\beta}$, $\hat{\theta}$, and $\hat{J_{3D}}$ are ground-truth annotations. Although the network can produce accurate body poses with the above constraints, the absolute positions may be totally unreasonable due to the depth-shape coupling. For example, a short person close to the camera can get a similar reprojection error as a tall person in the distance. Previous works~\cite{zanfir2018monocular,mehta2019xnect} rely on a known ground plane to decouple the ambiguities. However, the ground plane is not always available in a single in-the-wild crowd image. Thus, we further exploit the crowd cues and propose a crowd constraint to promote more accurate absolute position prediction. 
\begin{equation}
    \mathcal{L}_{\text{crowd}} = std(J_{root} \cdot l),\label{equ:crowd}
\end{equation}
where $std(\cdot)$ denotes standard deviation, and $J_{root} \in \mathbb{R}^{N \times 3}$ is the root positions of all people in the image. $(\cdot)$ means dot product.
\begin{equation}
l = \frac{1}{N} \sum_{n=1}^N \frac{J_{\text{top}}^{n} - J_{\text{bottom}}^{n}}{\|J_{\text{top}}^{n} - J_{\text{bottom}}^{n} \|}.
\end{equation}
$J_{\text{top}}$ is 3D keypoint on the head, and $J_{\text{bottom}}$ is the middle point of two ankle keypoints. The constraint penalizes unreasonable absolute positions and enforces more accurate body shapes. We found the constraint can be pretty robust in common crowded scenes with an appropriate loss weight.

%% file: dataset.tex
\section{Pseudo ground-truth for crowd data}\label{sec:Pseudodata}

Although 3D human data have seen prosperous developments in recent years, the crowd data in large-scale scenes is still scarce due to the requirement of complex hardware~\cite{wang2020panda} and expensive annotation~\cite{ionescu2013human3}. To promote the research on crowd analysis, recent works~\cite{fabbri2018learning,patel2021agora} produce photorealistic crowd data using game engines, rendering techniques or generative models~\cite{zhao2023semi}. However, a large domain gap exists between synthetic and real data since the illumination conditions and human textures are more complex in the real world. In addition, the natural human behavior in crowds can hardly be simulated in virtual environments.

Therefore, we follow the previous pseudo annotator~\cite{joo2021exemplar} to build 3D pseudo ground-truth~(GT) for  Panda~\cite{wang2020panda} and CrowdPose~\cite{li2019crowdpose}. CrowdPose is a crowd dataset for 2D pose estimation, and Panda is the first gigapixel-level human dataset, which captures real-world crowds ranging from 80 to 900 people. Since Panda does not contain 2D poses, we first predict the 2D keypoints with ViTPose~\cite{xu2022vitpose}. To generate 3D annotations, we first train our hypergraph relational reasoning network on common crowd data. Once the network is trained, we then estimate initial SMPL parameters from the 2D poses. Due to the domain gap, the initial values may not be accurate enough. We thus finetune the network parameters to adapt to 2D poses via reprojection error in \cref{equ:reproj}. Like \cite{joo2021exemplar}, we optimize the network parameters for several iterations and finally output the estimated results as the pseudo GT. Additional constraints are also used in the finetuning. The detailed procedure can be found in the Sup. Mat.

We also manually filter the incorrect estimations in the camera view. Different from previous pseudo annotators~\cite{joo2021exemplar,cliff}, the adaption explicitly considers the crowd interactions and constraints in multi-person scenarios. The 3D models in the final dataset have plausible ordinal relationships and are consistent with image observations. The experiment in \cref{tab:gigacrowd} shows that crowd reconstruction methods can gain significant improvement with the proposed datasets.

%% file: experiments.tex
\section{Experiments}\label{sec:experiment}

\begin{figure*}
    \begin{center}
    \includegraphics[width=1.0\linewidth]{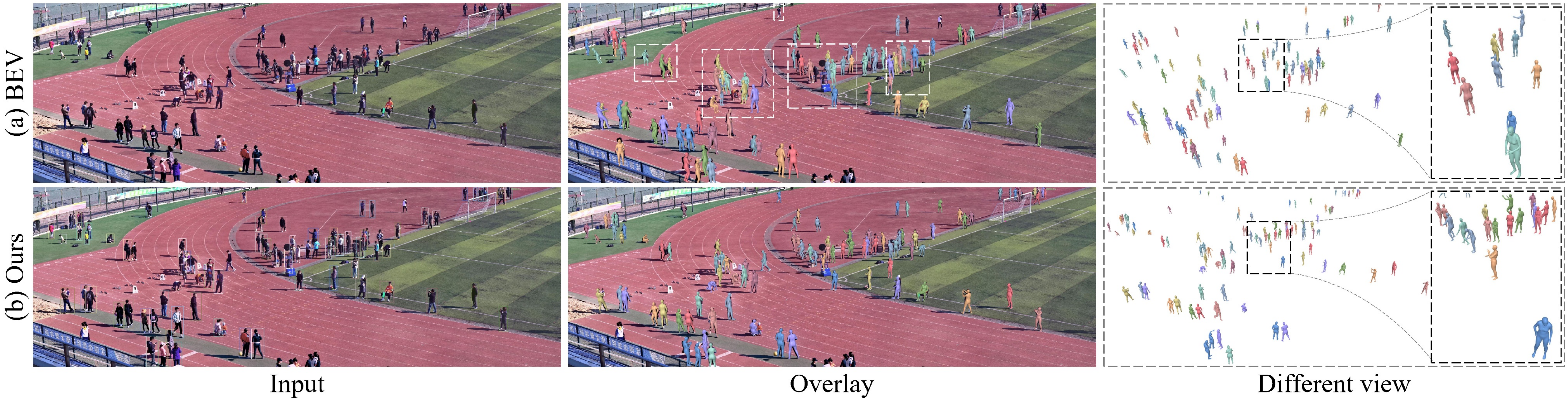}
    \end{center}
    \vspace{-7mm}
    \caption{Qualitative comparison with BEV~\cite{sun2022putting} on \textbf{GigaCrowd}. Our method is more robust to scale variations and occlusions. In addition, the proposed approach can also reconstruct crowds with more reasonable ordinal relationships.}
\label{fig:qualitative_comparison}
\vspace{-5mm}
\end{figure*}

\subsection{Datasets}\label{sec:datasets}
We use 3 benchmarks, Panoptic~\cite{joo2015panoptic}, GigaCrowd~\cite{gigacrowd}, and JTA~\cite{fabbri2018learning}, to evaluate our method. Panoptic is a multi-person dataset captured in an indoor environment, and we use it for evaluation only. We follow previous work~\cite{sun2022putting} to train the network on Human3.6M~\cite{ionescu2013human3}, MuCo-3DHP~\cite{mehta2018single}, MSCOCO~\cite{lin2014microsoft}, MPII~\cite{andriluka20142d} and CrowdPose~\cite{li2019crowdpose}. To further evaluate our method on more complex crowded scenes, we use GigaCrowd~\cite{gigacrowd}, a large-scale 3D crowd reconstruction dataset containing 3D root positions and 2D poses, as a benchmark. For the evaluation on GigaCrowd, besides the mentioned training data, the proposed Panda dataset is also used for training. On JTA dataset, we use its standard train and test split protocols to conduct the experiments. More details about each dataset can be found in Sup. Mat.

\subsection{Metrics}\label{sec:Metrics}
We follow \cite{wen2023crowd3d} to use the Procrustes-aligned pair-wise percentual distance similarity~(PA-PPDS)~\cite{wen2023crowd3d} and object keypoint similarity~(OKS)~\cite{lin2014microsoft} to evaluate the absolute positions and pose accuracy on GigaCrowd. In addition, the percentage of correct ordinal depth~(PCOD)~\cite{zhen2020smap} is adopted to measure the correctness of ordinal depth relations. The redundant punishment~(RP)~\cite{wen2023crowd3d} is also used to penalize redundant detections. On other datasets, we adopt the 3D extension of the Percentage of Correct Keypoints~($3DPCK$) and the Mean per Joint Position Error~(MPJPE) to measure the joint accuracy. To consider the missing detections, we follow \cite{cheng2022dual,fabbri2020compressed} to use F1-score with thresholds 0.4$m$, 0.8$m$, and 1.2$m$ for evaluating absolute positions. The detailed definition of metrics can be found in Sup. Mat.

\begin{table}
    \begin{center}
        \resizebox{1.0\linewidth}{!}{
            \begin{tabular}{l|c c c c}
            \noalign{\hrule height 1.5pt}
            \begin{tabular}[l]{l}\multirow{1}{*}{Method}\end{tabular}
                &PA-PPDS$\uparrow$ &OKS$\uparrow$ &PCOD$\uparrow$ &RP$\downarrow$  \\
            \noalign{\hrule height 1pt}
            \hline \hline
            % ~\etal~\cite{} & & & & & & \\
            CRMH~\cite{jiang2020coherent}         &63.29   &64.52        &75.28   &\textbf{0.17}            \\
            BEV~\cite{sun2022putting}             &71.37   &71.96        &83.27   &0.23             \\
            \textbf{Ours}                         &\textbf{82.21}   &\textbf{77.31}        &\textbf{88.21}   &\textbf{0.17}  \\
            \hline
            CRMH w/o Panda~\cite{jiang2020coherent}   &52.16   &56.31    &60.48      &\textbf{0.17}            \\
            BEV w/o Panda~\cite{sun2022putting}       &55.41   &62.47    &62.38   &0.22             \\
            \textbf{Ours} w/o Panda                   &\textbf{67.22}   &\textbf{70.80}    &\textbf{71.42}       &\textbf{0.17}  \\
            \noalign{\hrule height 1.5pt}
            \end{tabular}
        }
\vspace{-3mm}
\caption{\textbf{Comparisons on GigaCrowd}. "w/o Panda" means the model is trained without our Panda dataset.}
\label{tab:gigacrowd}
\end{center}
\vspace{-10.5mm}
\end{table}

\subsection{Comparison to state-of-the-art methods}\label{sec:Comparison}
We conduct several experiments to demonstrate the effectiveness of our method on large-scale crowded scenes. \cref{tab:gigacrowd} shows a quantitative comparison on GigaCrowd with CRMH and BEV. CRMH and BEV are the current SOTA methods that can obtain absolute body meshes in large-scale scenes. For a fair comparison, we train the baseline methods with the same data. Since rescaling the full image for BEV input causes extremely low resolution, we use BEV's released code to crop the original image and combine the predicted results from each patch. Due to the crowd constraints, our method can obtain more reasonable absolute positions and thus results in better performance in terms of PA-PPDS and PCOD. Since some crowds in GigaCrowd dataset show significant collectiveness, our method benefits from the group features and can outperform previous top-down and bottom-up approaches by a large margin on OKS. Besides, we found that the models trained on common multi-person data do not generalize well on large images due to severe scale variations and complex spatial distributions. The comparisons between rows 1-3 and rows 4-6 in \cref{tab:gigacrowd} reveal that the proposed Panda dataset can close the gap between common and large-scale scenarios. In \cref{fig:qualitative_comparison}, although BEV estimates the crowd from the cropped images, it still misses some people in the distance. Besides, BEV fails to estimate correct absolute positions in large-scale scenes, while our method produces a reasonable spatial distribution with the relational reasoning.

\begin{table}
    \begin{center}
        \resizebox{1.0\linewidth}{!}{
            \begin{tabular}{l|c c c c}
            \noalign{\hrule height 1.5pt}
            \begin{tabular}[l]{l}\multirow{1}{*}{Method}\end{tabular}
                &$3DPCK_{all}\uparrow$ &F1(0.4)$\uparrow$ &F1(0.8)$\uparrow$ &F1(1.2)$\uparrow$  \\
            \noalign{\hrule height 1pt}
            \hline \hline
            PandaNet~\cite{benzine2020pandanet}   &83.2        &--            &--            &--      \\
            Benzine~\etal~\cite{benzine2021single}&43.9        &--            &--            &--      \\
            LoCO~\cite{fabbri2020compressed}      &--           &50.82        &64.76        &70.44   \\
            Cheng~\etal~\cite{cheng2021monocular} &--           &57.22        &68.51        &72.86  \\ 
            Cheng~\etal~\cite{cheng2022dual}      &--           &58.15        &69.32        &74.19  \\ 
            \hline
            \textbf{Ours}                         &\textbf{86.7} &\textbf{59.59} &\textbf{70.81} &\textbf{76.67} \\
            \noalign{\hrule height 1.5pt}
            \end{tabular}
        }
\vspace{-3mm}
\caption{\textbf{Comparisons on JTA.} Due to the lack of SMPL annotations, we regress joint positions on this dataset for fair comparisons. Our method outperforms previous joint regression baseline methods. "--" means the results are not available.}
\label{tab:jta}
\end{center}
\vspace{-8mm}
\end{table}

\begin{table}
    \begin{center}
        \resizebox{1.0\linewidth}{!}{
            \begin{tabular}{l|c c c c c }
            \noalign{\hrule height 1.5pt}
            \begin{tabular}[l]{l}\multirow{1}{*}{Method}\end{tabular}
                &Haggling$\downarrow$ &Mafia$\downarrow$ &Ultim$\downarrow$ &Pizza$\downarrow$ &Mean$\downarrow$   \\
            \noalign{\hrule height 1pt}
    
            \hline \hline
    
            % ~\etal~\cite{} & & & & & & \\
            Zanfir~\etal~\cite{zanfir2018monocular}   &140.0 &165.9 &150.7 &156.0  &153.4        \\
            MubyNet~\cite{zanfir2018deep}       &141.4 &152.3 &145.0 &162.5  &150.3        \\
            % BMP-Lite~\cite{zhang2021body}       &124.2 &138.1 &155.2 &157.3  &143.7        \\
            CRMH~\cite{jiang2020coherent}       &129.6 &133.5 &153.0   &156.7  &143.2        \\
            BMP~\cite{zhang2021body}            &120.4 &132.7 &140.9 &147.5  &135.4        \\
            Pose2UV~\cite{huang2022pose2uv}     &104.2 &136.0   &123.2 &151.0    &128.6        \\
            ROMP~\cite{sun2021monocular}        &110.8 &122.8 &141.6 &137.6  &128.2        \\
            3DCrowdNet~\cite{choi2022learning}  &109.6 &135.9 &129.8 &135.6  &127.3        \\
            Luvizon~\etal~\cite{SceneAware_EG2023}  &93.6  &-- &133.8 &145.9  &--        \\
            BEV~\cite{sun2022putting}           &90.7  &\textbf{103.7} &113.1 &125.2  &109.5        \\
            
            \hline
            \textbf{Ours}                       &\textbf{86.8} &107.8 &\textbf{110.7} &\textbf{121.1} &\textbf{106.6} \\
            \noalign{\hrule height 1.5pt}
            \end{tabular}
        }
\vspace{-3mm}
\caption{Comparison with multi-person mesh recovery methods on \textbf{Panoptic} dataset. \textbf{All methods do not use the data from Panoptic for training.} The results for baseline methods are directly obtained from the original papers. The numbers are MPJPE.}
\label{tab:panoptic}
\end{center}
\vspace{-10.5mm}
\end{table}

With the group information and crowd constraints, our method also produces accurate body meshes and reasonable absolute positions on Internet images in \cref{fig:qualitative_results}. For the images that show significant collectiveness, the occluded people can obtain additional knowledge from others in the same group and result in appropriate poses.

We further conduct a comparison with previous pose estimation methods on JTA dataset in \cref{tab:jta}. PandaNet~\cite{benzine2020pandanet} is the first 3D pose estimation framework designed for large-scale scenes. Due to our top-down strategy, our method outperforms PandaNet in terms of 3DPCK. We also follow recent works~\cite{cheng2022dual,fabbri2020compressed} to use F1-score to validate the effectiveness of our method for absolute multi-person position prediction. For a fair comparison, we also regress 3D joint positions on this dataset. Our method constrains individuals with group features and still outperforms these works on all metrics.

To demonstrate the performance of our work on common multi-person scenarios, we compare existing multi-person mesh recovery works on Panoptic. Panoptic is captured in an indoor studio with designed activities, and the people in the image have a high pose similarity. Although the dataset has different camera views and severe mutual occlusions, our method still works well under this challenging setting. In \cref{tab:panoptic}, the Pizza sequence contains truncations, object and human occlusions, and our method also outperforms other baseline methods. The results show that the group features can compensate for insufficient individual information to address the occlusion, which reveals the importance of group features in multi-person mesh recovery.

\begin{figure}
    \begin{center}
    \includegraphics[width=1.0\linewidth]{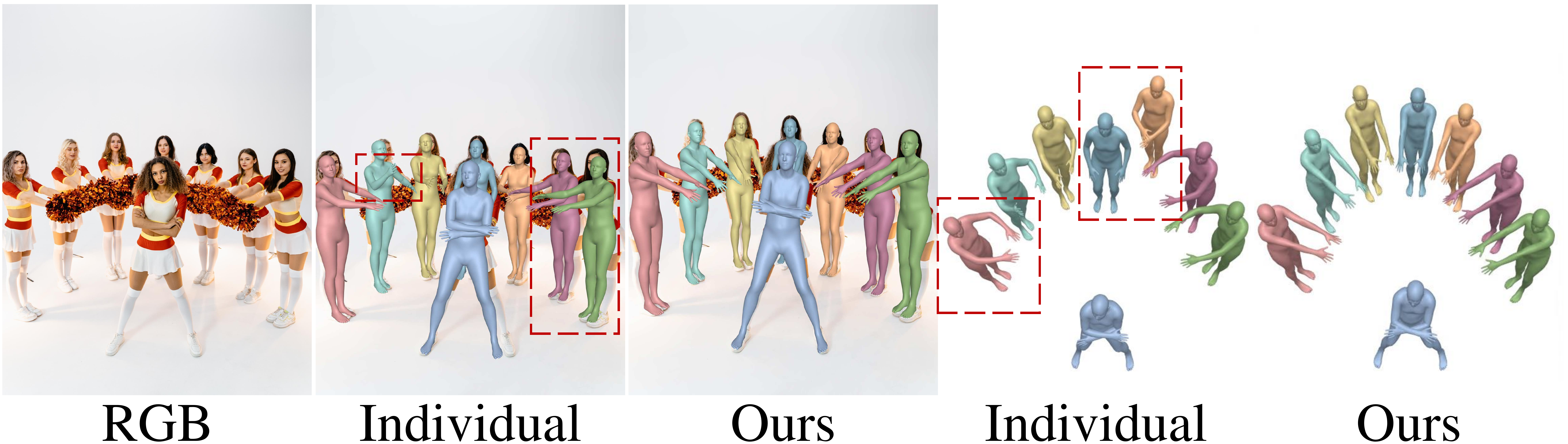}
    \end{center}
    \vspace{-6mm}
    \caption{"Individual" removes the relational reasoning. Our method can produce reasonable absolute positions and accurate body poses with the relational reasoning.}
\label{fig:ablation}
\vspace{-3mm}
\end{figure}

\begin{table}
    \begin{center}
        \resizebox{0.9\linewidth}{!}{
            \begin{tabular}{l|c c c c}
            \noalign{\hrule height 1.5pt}
            Method
                &CRMH~\cite{jiang2020coherent} &BEV~\cite{sun2022putting} &Individual &Ours  \\
            \noalign{\hrule height 1pt}
            \hline \hline
            % ~\etal~\cite{} & & & & & & \\
            FPS   &21.9        &24.1            &26.4   &23.2            \\
            \noalign{\hrule height 1.5pt}
            \end{tabular}
        }
\vspace{-3mm}
\caption{Running time on Panoptic with an RTX 3090 GPU. All top-down methods use YOLOX~\cite{yolox2021} for detection.}
\label{tab:complexity}
\end{center}
\vspace{-10mm}
\end{table}

\subsection{Ablation study}\label{sec:Ablation}

\textbf{Relational reasoning.} We investigate the importance of group features for multi-person mesh recovery. In \cref{tab:ablation}~(Individual), we adopt the network without relational reasoning and directly regress human meshes from individual features via the head network in \cref{sec:regression}, which shows a significant decrease. In \cref{fig:ablation}, all individuals are improved from the collectiveness with the relational reasoning. Then, we replace the proposed hypergraph network with a transformer for the relational reasoning~(Transformer). The transformer-based network is similar to \cite{kimmulti}, which receives individual features for $N$ people and outputs corresponding SMPL parameters. It implicitly learns human correlations with attention mechanisms and ignores group-wise relations. Conversely, our hypergraph relational reasoning explicitly forms groups with crowd collectiveness and considers the group behavior’s influence, which leads to superior performance.

\textbf{Group size and scales.} We analyze the impact of group size and scales in \cref{tab:ablation}. The performance increases with more scales at first and then becomes stable. In addition, the people in large groups ~(\eg, 11) in most cases have unobvious crowd collectiveness, and the group information may introduce noises in the reasoning. 

\textbf{Crowd constraints.} Due to the depth ambiguity, regressing reasonable absolute positions from monocular image is an ill-posed problem. The PA-PPDS in \cref{tab:ablation} shows that the ambiguity can be greatly alleviated by incorporating the crowd constraints in the loss function.

\textbf{Computational complexity.} We compare the running time and network parameters in \cref{tab:complexity} and \cref{tab:ablation}. The results show that the relational reasoning is compact, and our method has competitive running efficiency.

\begin{table}
    \begin{center}
        \resizebox{1.0\linewidth}{!}{
            \begin{tabular}{l|c c c}
            \noalign{\hrule height 1.5pt}
            \begin{tabular}[l]{l}\multirow{1}{*}{Method}\end{tabular}
                &PA-PPDS$\uparrow$ &OKS$\uparrow$ &Params$\downarrow$  \\
            \noalign{\hrule height 1pt}
            \hline \hline
            % ~\etal~\cite{} & & & & & & \\
            Individual             &70.20         &69.44             &26.47M               \\
            Transformer            &76.47         &70.28             &29.24M                \\
            hypergraph-(1)         &75.71         &71.06             &27.15M               \\
            hypergraph-(1,3)       &78.11         &73.30             &29.17M                \\
            hypergraph-(1,3,5)     &82.21         &77.31             &29.18M                  \\
            hypergraph-(1,2,3,5)   &82.61         &76.78             &30.19M                 \\
            hypergraph-(1,3,5,11)  &81.84         &75.41             &30.19M                 \\
            hypergraph-(1,3,5) w/o $\mathcal{L}_{\text{crowd}}$    &77.07         &73.41             &29.18M                 \\
            \noalign{\hrule height 1.5pt}
            \end{tabular}
        }
\vspace{-3mm}
\caption{\textbf{Ablation studies on GigaCrowd.} "Transformer" uses a transformer-based network for relational reasoning. "(1,3,5)" means 3 scales with group sizes of 1, 3, and 5.}
\label{tab:ablation}
\end{center}
\vspace{-10mm}
\end{table}

%% file: conclusion.tex
\section{Limitation and future work}\label{sec:Limitation}
Although our method can reconstruct human groups in large-scale crowd images, there still exist some limitations. First, when the number of people in an image exceeds the maximum, the relational reasoning can only afford a limited number of individuals at a time. Although we can still simultaneously estimate all people in the image by assigning them to different samples of a batch, an interactive pair may be assigned to different samples and can not provide additional cues for each other. In the future, the network can be improved to aggregate similar body poses in the same node to get better compatibility. Second, we may require to decrease the crowd constraint loss weight for some special cases where people are in different planes~(\eg, audience in a theater). A too-large crowd loss weight may drag the people to the same plane. To address this problem, to incorporate the scene semantics for future crowd reconstruction might be a feasible solution.

\section{Conclusion}\label{sec:Conclusion}
In this work, we propose a novel hypergraph relational reasoning network to exploit crowd features for reconstructing groups of people from a large-scale monocular image. To construct the graph topology, crowd collectiveness is used to infer the connection relationships. The proposed network explicitly considers both pair-wise and group-wise relations with a compact individual representation, and promotes accurate body pose and absolute position prediction. In addition, we also build pseudo ground-truth for two crowd datasets. The proposed datasets may promote future research on pose estimation and human behavior understanding in crowded scenes.